\documentclass{article}

     \PassOptionsToPackage{numbers}{natbib}

 \usepackage[preprint]{neurips_2019}

\usepackage[utf8]{inputenc} 
\usepackage[T1]{fontenc}    
\usepackage{hyperref}       
\usepackage{url}            
\usepackage{booktabs}       
\usepackage{amsfonts}       
\usepackage{nicefrac}       
\usepackage{microtype}      
\usepackage{bm}
\usepackage{graphicx}
\usepackage[ruled]{algorithm2e}
\usepackage{mathtools,amssymb}
\usepackage{multirow}
\usepackage{booktabs}

\title{Epoch-evolving Gaussian Process Guided Learning}

\author{%
  Jiabao Cui,~~Xuewei~Li,~~Bin~Li,~~Hanbin~Zhao,~~Bourahla~Omar,~~Xi~Li  \\
	\texttt{Zhejiang University}\\
  \texttt{\{jbcui,3150104097,binli,zhaohanbin,bourahla,xilizju\}@zju.edu.cn} \\
}
\begin{document}

\maketitle

\begin{abstract}
In this paper, we propose a novel learning scheme called epoch-evolving Gaussian Process Guided Learning (GPGL), which aims at characterizing the correlation information between the batch-level distribution and the global data distribution. Such correlation information is encoded as context labels and needs renewal every epoch. With the guidance of the context label and ground truth label, GPGL scheme provides a more efficient optimization through updating the model parameters with a triangle consistency loss. Furthermore, our GPGL scheme can be further generalized and naturally applied to the current deep models, outperforming the existing batch-based state-of-the-art models on mainstream datasets (CIFAR-10, CIFAR-100, and Tiny-ImageNet) remarkably.

\end{abstract}

\section{Introduction}

Recent years have witnessed a great development of deep learning with a wide range of applications. 
Due to the computational resource limit, it has to rely on the mini-batch stochastic gradient descent (SGD~\cite{bottou1998online}) algorithm for iterative model learning over a sequence of epochs, each of which corresponds to a collection of randomly generated sample batches with small sizes. 
In the learning process, it asynchronously updates the model parameters with respect to time-varying sample batches, and thus captures the local batch-level distribution information, resulting in the ``zig-zag'' effect in the optimization process~\cite{qian1999momentum}. 
Therefore, it usually requires a large number of epoch iterations for sufficient model learning, which essentially takes a bottom-up learning pipeline from local batches to global data distribution. Such a pipeline is incapable of effectively balancing the correlation information between batch-level distribution and global data distribution for those sequentially-added sample batches within different epochs.     

Motivated by the above observation, we seek for a possible solution to enhance the bottom-up learning pipeline with the top-down strategy, which aims at approximately encoding the global data distribution information as class distribution~\cite{laurikkala2001improving} by a nonparametric learning model~\cite{roy2018bayesian}. 
As a result, we have a hybrid learning pipeline that effectively combines the benefits of precise batch-level learning and global distribution-aware nonparametric modeling. 
More specifically, we propose an epoch-evolving Gaussian Process Guided Learning (GPGL) scheme that dynamically estimates the class distribution information for any sample with the evolution of epochs in a nonparametric learning manner.
For each epoch learning, the proposed Gaussian process approach builds a class distribution regression model in the corresponding epoch-related feature space to estimate the class distribution as context label for any sample, relative to a set of fixed class-aware anchor samples. 
In essence, this context label estimation corresponds to a contextual label propagation process, where the class distribution information from the class-aware anchor samples are dynamically propagated to the given samples through Gaussian process regression~\cite{Rasmussen:2005:GPM:1162254}. 
Subsequently, with the guidance of the propagated context label, the deep model can learn the class distribution information in the conventional learning pipeline.
Hence, we have a triangle consistency loss function consisting of three learning components: 1) deep model prediction with ground truth label; 2) deep model prediction with context label; and 3) context label with ground truth label.
The triangle consistency loss function is jointly optimized for each epoch.
After one epoch, the epoch-related feature space is accordingly updated from the latest deep model. Based on the updated feature space, the joint triangle consistency loss is optimized once again in the next epoch.
The above learning process is repeated until convergence.

In principle, the epoch-evolving GPGL scheme takes into account the context dependency relationships between given samples and class-aware anchor samples, which effectively explores global data distribution in a nonparametric modeling fashion. 
Such a distribution structure usually carries a rich body of contextual information that is capable of alleviating the ``zig-zag'' problem and meanwhile speeding up the convergence process. 
The joint triangle consistency loss seeks for a good balance of deep learning features, deep learning prediction, Gaussian process context label, and ground truth for each sample. 
In summary, the main contributions of this work are as follows:
\begin{itemize}
	\setlength{\itemsep}{0pt}
	\setlength{\parsep}{0pt}
	\setlength{\parskip}{0pt}
	
	\item We propose an epoch-evolving GPGL scheme that takes a nonparametric modeling strategy for estimating the context-aware class distribution information to guide the model learning process. Based on the Gaussian process approach, we set up a hybrid bottom-up and top-down learning pipeline for effective model learning with better convergence performance.
	\item We present a joint triangle consistency loss function that is capable of achieving a good balance between batch-level learning and global distribution-aware nonparametric modeling. Experimental results over the benchmark datasets show this work achieves the state-of-the-art results. 
\end{itemize}

\paragraph{Related work}
Gaussian Processes (GPs)~\cite{Rasmussen:2005:GPM:1162254} are a class of powerful, useful, and flexible Bayesian nonparametric probabilistic models. With the rapid development of deep learning, GPs have been generalized with multi-layer neural network as Deep Gaussian Processes (DGPs)~\cite{damianou2013deep, bui2016deep}. In DGPs, the mapping between layers is parameterized as GPs such that the uncertainty of prediction could be accurately estimated. On the other hand, variational inference, which aims at approximating the posterior distribution over the latent variables, is introduced in to augment deep Gaussian processes to form Variational Gaussian Processes (VGPs)~\cite{tran2015variational, dai2015variational, casale2018gaussian,li2011graph}. However, both DGPs and VGPs are focusing on feature space, our work mainly apply GPs to seek a more ideal label space.


Optimization of deep learning model mainly focuses on the training algorithms and hyper-parameter tuning~\cite{he2016deep,huang2017densely,he2016identity,zagoruyko2016wide,summaira2021recent,jiang2019learning, chen2014ranking}. Roughly speaking, two classes of methods are commonly categorized. The first class, e.g., SGD Momentum  (SGD-M~\cite{sutskever2013importance}), Nesterov~\cite{nesterov1983method}, utilizes the momentum to fix the current update direction. The other class designs an adaptive learning rate adjustment scheme, such as AdaGrad~\cite{duchi2011adaptive}, RMSProp~\cite{hinton2012lecture}, and Adam~\cite{kingma2014adam}. 
However, we argue that these conventional training approaches focus on batch-level data distribution in optimization process without the attention of global data distribution information. Motivated by this, we propose the epoch-evolving GPGL to alleviate this concern by introducing the contextual information of the dataset.

\section{Method}
\label{Method}
\subsection{Overview}
\label{Overview}
In this section, we describe the overview of the epoch-evolving Gaussian Process Guided Learning (GPGL) scheme. Given a dataset $\mathcal{D}=\{(\mathbf{x}^{i}, y^{i})\}_{i=1}^{N}$, $\mathbf{x}^{i} \in \mathcal{X}$ is the input data and $y^{i} \in \mathcal{Y} = \{1, 2, ..., C\}$ is the corresponding label. Let $\mathbf{h}^{i} = f(\mathbf{x}^{i})$ be the feature extracted by function $f$ and $\hat{y}^{i}=g(\mathbf{h}^{i})$ be the predicted label by function $g$. Since the conventional bottom-up deep learning pipeline usually applies a batch-based optimization paradigm, we denote $\hat{y}_b$, ${y}_b$, $\mathbf{h}_b$ as the prediction, ground truth, and feature of a single sample $\mathbf{x}_b$ in a batch $\mathcal{B}$. The routine optimization framework is interpreted as:

\begin{equation}\label{mBSGD_update}
\omega_{t+1} = \omega_t - 
\frac{\eta}{|\mathcal{B}|} \sum_{b \in  \mathcal{B}}^{|\mathcal{B}|}\nabla\mathcal{L}
\left(\hat{y}_b ,{y}_b\right),
\end{equation}
where a batch of $|\mathcal{B}|$ samples is randomly picked and optimized with learning rate $\eta$, and $\mathcal{L}$ is the classification loss function.

The function $g$ in conventional DNN setting is determined by its weight $\omega$ uniquely. Usually, we apply a stochastic gradient optimization algorithm based on the mini-batch in each iteration. The specific batch of samples, however, only contains a relatively limited amount of data compared with the whole dataset. Hence, the update of weight $\omega$ is easily governed by the current batch, resulting in ``zig-zag'' effect. In other words, we argue that the conventional batch-based optimization paradigm can easily be affected by current batch without respect to the global data distribution information.

\begin{figure*}[!tp]
	\centering
	\makebox[\textwidth][c]{\includegraphics[scale=1.1]{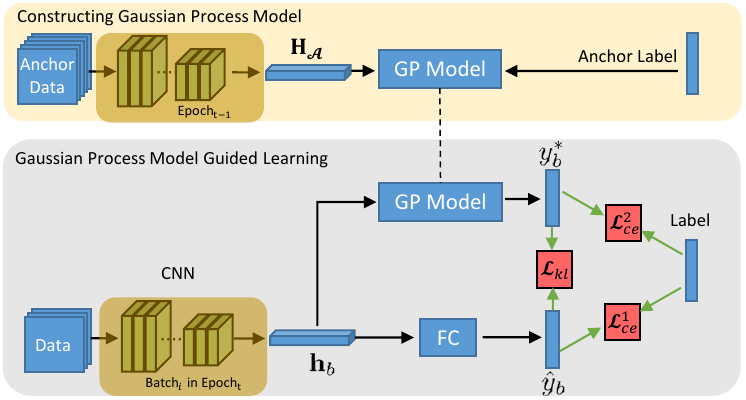}}
	\caption{Overview of Gaussian Process Guided Learning (GPGL) scheme. The GP model leverages the features and labels of the anchor set to construct the global data distribution approximately. The architecture of the feature extractor is the same as that of the deep model in batch-level learning process. However, the parameters of feature extractor in GP model is updated at the epoch-level. After constructing the GP model, it would make a context label $y^*_b$ for each sample, and guide the deep model learning by the triangle loss function. }
	\label{Method_1}
\end{figure*}

Motivated by the above challenge, we explore the class distribution predictions which encode the global data distribution. Inspired by the conventional nonparametric generative model, we propose a GPGL scheme to make class distribution predictions which we name as the context label. As shown in Figure~\ref{Method_1}, our GPGL scheme consists of Gaussian process model construction and Gaussian process model guided learning. Based on the observation data, our Gaussian process model builds the joint probability distribution which could embed the global distribution of the data in a nonparametric way. Therefore, the distribution of batch data is well fixed by the guidance of GP model and thus highly mitigates the effect of the aforementioned concern. A question then occurs on how to encode the global data distribution with our GP model.

We firstly extract a subset $\mathbf{X}_{\mathcal{A}} = \{\mathbf{x}^{i}\}_{i \in \mathcal{A}}$ and $\mathbf{y}_{\mathcal{A}} = \{{y}^{i}\}_{i \in \mathcal{A}}$ called the anchor set as a representative of the whole dataset due to the limited computational resources. The feature of the anchor set is then defined as $\mathbf{H}_{\mathcal{A}} = f(\mathbf{X}_{\mathcal{A}})$. A Gaussian process (GP) is thus build upon both anchor set and a coming sample:

\begin{equation}
\label{GP_joint_dis}
\begin{pmatrix*}[l]
\mathbf{y}_{\mathcal{A}} \\
{y}_{b}^{*} \\
\end{pmatrix*}
\sim \mathcal{N}\left(
\mathbf{0}, 
\begin{bmatrix*}[l]
K(\mathbf{H}_{\mathcal{A}}, \mathbf{H}_{\mathcal{A}}) &  
K(\mathbf{H}_{\mathcal{A}}, \mathbf{h}_{b})\\
K(\mathbf{h}_{b}, \mathbf{H}_{\mathcal{A}}) &  
K(\mathbf{h}_{b}, \mathbf{h}_{b})\\
\end{bmatrix*}
\right),
\end{equation}

where ${y}_{b}^{*}$ denotes the context label of a sample which embeds the whole dataset with annotation and $K(\cdot, \cdot)$ is the covariance function. The context label will further be applied into our novel triangle loss. In the following section, we would describe our context label construction and triangle loss in detail.

\subsection{Context Label construction}
\label{Context_label_construction}

The one-hot vector is commonly used as supervision in conventional work for its simplicity. However, since labels are encoded as discrete form, every two different labels are orthogonal to each other, neglecting the relationship among label pairs. We respectfully point that the label itself represents a space. Thus a perfect label space should hold both smoothness and consistency with feature space. To make it clear, if two samples are similar in feature space, the same similarity should hold in label space.

From the perspective of statistical learning, we seek an ideal label space which fulfills the above requirements based on the one-hot label distribution. The Gaussian process, which assumes any collection of its random variables satisfies a joint multivariate distribution, is thus applied to obtain an ideal label space. Given a label prior in the one-hot format, our method shows two advantages. On the one hand, we infer a better form of label which we term as a context label. On the other hand, the context label could better guide the representation learning. 

The context label is estimated by our Gaussian process model with the help of the anchor set described in the Section~\ref{Overview}. The anchor set is an abstraction of the overall dataset which contains the contextual information between different classes.
How to make a context label with respect to the correlation among a sample and the anchor set samples is a big challenge. 
Bayesian prediction is a powerful nonparametric method to make a posterior prediction based on the prior distribution. 
Rethinking the conventional batch-by-batch learning pipeline, any finite collection of sample's context label ${y}_{b}^{*}$ and anchor set labels $\mathbf{y}_{\mathcal{A}}$ follow the joint multivariate Gaussian distribution. 
In this general assumption, we make a Bayesian prediction for any coming sample $\mathbf{x}_b$ (with feature $\mathbf{h}_{b} $) in batch data based on the features $\mathbf{H}_{\mathcal{A}}$ and one-hot labels $\mathbf{y}_{\mathcal{A}}$ of the anchor set $\mathbf{X}_{\mathcal{A}}$.

Following the above notation, $\mathbf{y}_{\mathcal{A}}, {y}_{b}^{*} | \mathbf{H}_{\mathcal{A}}, \mathbf{h}_{b} \sim \mathcal{N}({\mathbf{0}, \mathbf{K}})$ is a multivariate normal distribution whose covariance matrix $\mathbf{K}$ has the same form as Equation~(\ref{GP_joint_dis}). As mentioned above, the similarities of samples in feature space hold consistency with that of samples in smooth label space. How do we measure the distance between any two samples in feature space? Our Gaussian process model selects the RBF kernel to capture the similarity between different features:
\begin{equation}\label{RBF}
K(\mathbf{H}_{\mathcal{A}}, \mathbf{h}_{b})
= \mathrm{exp}\left\{
-\frac{r^2(\mathbf{H}_{\mathcal{A}}, \mathbf{h}_{b})}
{2l^2}
\right\},
\end{equation}

where $l$ is a length-scale parameter~\cite{jylanki2011robust} and $r^2\left(\cdot,\cdot\right)$ is the Euclidean distance function.

With this distance measurement, our Gaussian process model builds the aforementioned context label for any sample. The context label corresponds to the correlation information among a sample and the anchor set samples. In our Gaussian process model, the label information of anchor samples is propagated by a Gaussian process regression (GPR)~\cite{Rasmussen:2005:GPM:1162254} approach, assuring the smoothness property in label space through interpolation. 
Therefore, the conditional distribution of context label can be done analytically ${y}_{b}^{*} | \mathbf{y}_{\mathcal{A}}, \mathbf{H}_{\mathcal{A}}, \mathbf{h}_{b} \sim \mathcal{N}(\mathbf{g}_m, {g}_v)$ with:

\begin{align}\label{G_M}
\mathbf{g}_m &= 
K(\mathbf{h}_{b}, \mathbf{H}_{\mathcal{A}})
\left(
K(\mathbf{H}_{\mathcal{A}}, \mathbf{H}_{\mathcal{A}}) + \sigma^2\mathbb{I}
\right)^{-1}
\mathbf{y}_{\mathcal{A}};\\
{g}_v &= 
K(\mathbf{h}_{b}, \mathbf{h}_{b}) -
K(\mathbf{h}_{b}, \mathbf{H}_{\mathcal{A}})
\left(
K(\mathbf{H}_{\mathcal{A}}, \mathbf{H}_{\mathcal{A}}) + \sigma^2\mathbb{I}
\right)^{-1}
K(\mathbf{H}_{\mathcal{A}}, \mathbf{h}_{b}) \label{G_V},
\end{align}

where $\mathbb{I}$ is the $n\times n$ identity matrix. The mean of distribution $\mathbf{g}_m$ represents our context label and the covariance function  $g_v$ controls to what extent we trust our context label.

\begin{figure*}[!tp]
	\centering
	\makebox[\textwidth][c]{\includegraphics[scale=0.77]{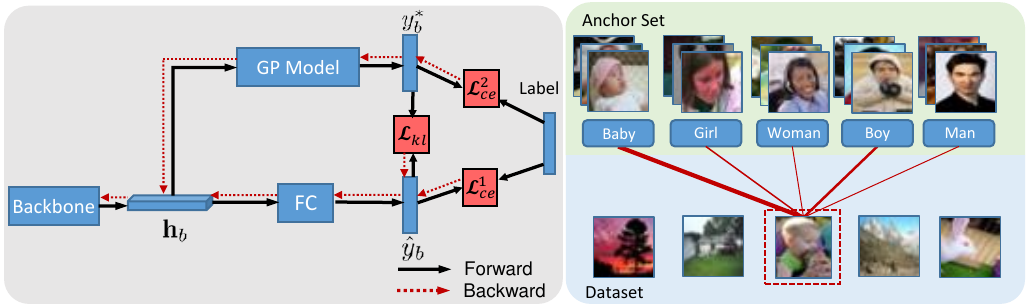}}
	\leftline{\small\hspace{1.5cm}(a) Triangle loss backward \hspace{4.0cm}(b) Class-aware anchor sample}
	\caption{(a) As shown in red dashed lines, the loss functions $\mathcal{L}_{ce}^1$ and $\mathcal{L}_{kl}$ are used for updating all the parameters in the deep neural network model, while the $\mathcal{L}_{ce}^2$ is used for updating the parameters of feature extractor. (b) The context label of an image in dataset is estimated by the correlation with each class in the anchor set. The width of red solid lines indicates the probability value of the context label for each class. Wider lines show a stronger correlation.}
	\label{method_2}
\end{figure*}

In our Gaussian process model, the inference time complexity for each sample is $\mathcal{O}\left( |\mathcal{A}|^2\right)$, where $|\mathcal{A}|$ is the size of the anchor set, which is infeasible in practice for a large anchor set. We provided a class-aware anchor sampling mechanism for picking the near $C_{cor}$ neighbors for each class by measuring the similarity between mean features of each class in feature space. With this mechanism, the anchor set could represent the global structure of data distribution approximately. The correlation between a sample and the anchor set samples is visualized in Figure~\ref{method_2}(b). 

\subsection{Triangle consistency loss}
\label{Triangle_loss}

\paragraph{The loss terms} With our Gaussian process (GP) model employed, there are two prediction terms for a sample $\mathbf{x}_b$: the deep model prediction $\hat{y}_{b}$ and the context label ${y}^{*}_{b}$. With the ground truth label $ y_b$, the conventional cross-entropy loss function $\mathcal{L}_{ce}^1 \left(\hat{y}_{b},y_b\right) $ between the model prediction and ground truth for a sample is defined as:
\begin{equation}\label{CE_1}
\mathcal{L}_{ce}^1 \left(\hat{y}_{b},y_b\right)  
= - \sum_{j=1}^{C} y_b^j \log(\hat{y}_b^j),
\end{equation}
where $\hat{y}_b^j$ is the model prediction for class $j$, and $y_b^j$ is the ground truth label for class $j$.
In addition to the conventional cross-entropy loss, we employ a KL divergence loss $\mathcal{L}_{kl} \left({y}^{*}_{b},\hat{y}_{b}\right)$ which is defined in Equation~(\ref{KL}) between the context label and deep model prediction. This loss term aims to reduce the difference between the prediction of the deep model and the prediction of our GP model. The KL divergence loss is used as a regularization term to prevent the deep model learning from overfitting.
\begin{equation}\label{KL}
\mathcal{L}_{kl} \left({y}^{*}_{b},\hat{y}_{b}\right)  
=- \sum_{j=1}^{C} {y}^{*j}_{b} \log(\frac{{y}^{*j}_{b}}{\hat{y}_b^j}),
\end{equation}
where ${y}^{*j}_{b}$ is the value of context label for class $j$. 
Our GP model produces a context label based on the feature space, and when the model is not well trained, the feature space is easily affected by any input data. So we use an extra loss term to constrain the transformation from data into feature space. We use a cross-entropy loss $\mathcal{L}_{ce}^2 \left({y}^{*}_{b},y_b\right)$  defined in Equation~(\ref{CE_2}) between context label and the ground truth. The backpropagation of this loss only flows through the feature extractor. Trying to minimize this loss term, the model can update the feature extractor's parameters, which improves the feature representation.

\begin{equation}\label{CE_2}
\mathcal{L}_{ce}^2 \left({y}^{*}_{b},y_b\right)  
= - \sum_{j=1}^{C} y_b^j \log({y}^{*j}_{b}).
\end{equation}

Our Gaussian process model is a nonparametric model based on the feature space. As indicated in Equation~(\ref{G_M}) and Equation~(\ref{G_V}), we build our Gaussian process model by Gaussian process regression. The backpropagation of loss function $\mathcal{L}_{ce}^2 $ only affects the convolutional layers which are used for the feature extractor. Both of the loss function $\mathcal{L}_{ce}^2 $ and  $\mathcal{L}_{kl} $ act on all the model parameters including convolutional layers and fully connected layers in backpropagation, as shown in Figure~\ref{method_2}(a).

\paragraph{Triangle consistency} With the above three terms of loss functions, there appears a triangle loss function between the deep model prediction, context label and ground truth label. The ground truth label is just one-hot for input data which cannot reflect the correlation information with other classes or samples. The context label is generated by our Gaussian process model which embeds the global data distribution. As our Gaussian process model is based on the feature space which is dynamically changing through learning process, the context label should be constrained by the ground truth to seek for better feature expressions. That is why we use a triangle consistency loss function among them. By introducing the triangle loss function, our GPGL scheme updates the parameters of the deep model with Equation~(\ref{mBSGD_update_loss_three}) in comparison to the conventional optimization framework with Equation~(\ref{mBSGD_update}).

\begin{equation}\label{mBSGD_update_loss_three}
\omega_{t+1} = \omega_t - 
\frac{\eta}{|\mathcal{B}|}\sum_{b=1}^{|\mathcal{B}|} \left( \nabla \alpha \mathcal{L}_{ce}^1 \left(\hat{y}_{b},y_b\right) + \nabla \beta \mathcal{L}_{kl} \left({y}^{*}_{b},\hat{y}_{b}\right) + \nabla \gamma \mathcal{L}_{ce}^2 \left({y}^{*}_{b},y_b\right)  \right),
\end{equation}
where $\alpha=\frac{1}{2-\mu}$, $\beta=\frac{|\mathcal{L}_{ce}^1|\left(1-\mu\right)}{2|\mathcal{L}_{ce}^2|\left(2-\mu\right)(1+g_v)}$, and $\gamma=\frac{|\mathcal{L}_{ce}^1|\left(1-\mu\right)}{2|\mathcal{L}_{kl}|\left(2-\mu\right)(1+g_v)}$ are the normalization items to balance the weight between the three loss terms. $\mu$ is the error rate of deep model initialized with random guess probability. $|\cdot|$ denotes the absolute value operator.

Rethinking the optimization framework with epoch sequences, for each epoch, all samples over the dataset are used once to update the model parameters, which makes the transformation of the data into a feature space changeable. Our Gaussian process model is based on the feature space and the features of the anchor set are updated after each epoch, hence our GP model evolves and is further utilized in next epoch (namely, epoch-evolving GP model). The optimization method of our proposed epoch-evolving GPGL scheme is shown in Algorithm~\ref{algo_1}. 


\begin{algorithm}[!h]
	\caption{Epoch-evolving Gaussian Process Guided Learning (GPGL)} \label{algo_1}
	\LinesNumbered
	\SetAlgoNoLine
	\KwIn{Training set $\mathcal{D}=\{(\mathbf{x}^{i}, y^{i})\}_{i=1}^{N}$ with $C$ classes; Training epochs $\tau$; Training batch size $|\mathcal{B}|$}
	Initialize (parameters $\omega$; error rate $\mu=1-\frac{1}{C}$;  $|\mathcal{L}_{ce}^1|= |\mathcal{L}_{ce}^2|=1$ (referred to Section~\ref{Triangle_loss}));\\
	Uniformly sample an anchor set $\mathbf{X}_{\mathcal{A}} = \{\mathbf{x}^{i}\}_{i \in \mathcal{A}}$ (referred to Section~\ref{Context_label_construction});\\
	Build our Gaussian process (GP) model based on the feature space of anchor set $\mathbf{X}_{\mathcal{A}}$;\\
	\For {\emph{epoch = $1, \cdots, \tau$} }{
		{
			
			\For {\emph{batch $ = 1, \cdots,$ $\frac{N}{|\mathcal{B}|}$ } }{
				Load and normalize the $|\mathcal{B}|$ samples;\\
				Compute the deep model prediction $\hat{y}_{b}$ (referred to Section~\ref{Overview});\\
				Compute the context label ${{y}^{*}_{b}}= \mathbf{g}_m $ with our GP model (Equation~(\ref{G_M}));\\
				Compute the loss functions $\mathcal{L}_{ce}^1 $, $\mathcal{L}_{kl}  $ and $\mathcal{L}_{ce}^2 $ (Equations~(\ref{CE_1}),~(\ref{KL}) and (\ref{CE_2}));\\
				Update the parameters $\omega$ (Equation~(\ref{mBSGD_update_loss_three}));
			}
			Evaluate the model and update $\mu$, $|\mathcal{L}_{ce}^1|$ and $|\mathcal{L}_{ce}^2|$;\\
			Update our GP model based on the new feature space; 
		}
	}
	\KwOut{Trained neural network model parameters $\omega$}
\end{algorithm}

\section{Experiments}

\subsection{Datasets}
\label{datasets}
\paragraph{CIFAR-10} It is a labeled subset of the 80 million tiny images dataset for object recognition. This dataset contains 60000 32$\times$32 RGB images in 10 classes, with 5000 images per class for training and 1000 images per class for testing.

\paragraph{CIFAR-100} It is a similar dataset to CIFAR-10 in that it also contains 60000 images, but it covers 100 fine-grained classes. Each class has 500 training$/$100 test images. 

\paragraph{Tiny-ImageNet} It has 200 different classes. 500 training images, 50 validation images, and 50 test images are contained in each class. Compared with CIFAR, the Tiny-ImageNet is more difficult because it has more classes and the target objects often cover just a tiny area in the image.

\subsection{Implementation details}
\label{Implement_datails}
\paragraph{Data preprocessing}
On CIFAR-10 and CIFAR-100, our data preprocessing is the same in~\cite{he2016deep}: 4 pixels are padded on each side, a 32$\times$32 crop is randomly sampled from the padded image or its horizontal flip. For testing, we only evaluate the single view of the original 32$\times$32 image.
On Tiny ImageNet, we use original images or their horizontal flip, rescale it to 256$\times$256, and sample 224$\times$224 crop randomly from the 256$\times$256 images. In testing, we only rescale images to 256$\times$256, and sample 224$\times$224 center crops.

\paragraph{Training details}
We use the same hyper-parameters as we replicate the others' results with SGD-Momentum (SGD-M) strategy when we train our model with GPGL scheme. In particular, we set the momentum to 0.9 and set the weight decay to 0.0001, following a standard learning rate schedule that goes from 0.1 to 0.01 at 60$\%$ (CIFAR-10)$/$50$\%$ (CIFAR-100)$/$33.3$\%$ (Tiny ImageNet) and to 0.001 at 80$\%$ (CIFAR-10)$/$75$\%$ (CIFAR-100)$/$66.7$\%$ (Tiny ImageNet) training. We use 250$/$200$/$120 epochs on CIFAR-10$/$CIFAR-100$/$Tiny ImageNet respectively. In addition, we train our models from random initialization on CIFAR-10$/$CIFAR-100. However, we load Imagenet pre-trained model and train the fully connected layer 10 epochs (learning rate = 0.1) on Tiny ImageNet dataset. In addition, we train our models on a server with 4 NVIDIA GTX 1080Ti. 

\subsection{Ablation experiments}
\label{Ablation_exp}

\paragraph{Setting of key hyper-parameters}
There are several key hyper-parameters in our GPGL: the length-scale parameter $l$ in the kernel function (mentioned in Equation~(\ref{RBF})), the number of classes included in the anchor set and the number of samples in each anchor class.
We observe the effectiveness of context labels with respect to the number of samples in each class, and find out that the performance of deep model is stable enough when the length-scale parameter $l$ and the number of samples in each anchor class are in a relatively large scale.
Therefore, we choose $l = $ 200$/$70$/$70 in the CIFAR-10$/$CIFAR-100$/$Tiny-ImageNet experiments for simplicity. For the number of samples, we use 128$/$7000$/$14000 on CIFAR-10$/$CIFAR-100$/$Tiny-ImageNet in practice. We calculate the mean of context labels for the whole CIFAR-100 test datasets and find that the ``Top-5'' classes contain about 70$\%$ information of the whole context label. ``Top-5'' classes are the 5 most related classes which are the 5 biggest values in the mean of context labels. When we do the comparative experiments according to the ``Top-5''$/$``Top-10''$/$``Top-20''$/$``Top-100'' strategies, we do not find any remarkable difference between them. We follow the ``Top-5'' strategy in our following experiments for simplicity.

\begin{figure*}[!tp]
	\centering
	\makebox[\textwidth][c]{\includegraphics[scale=0.5]{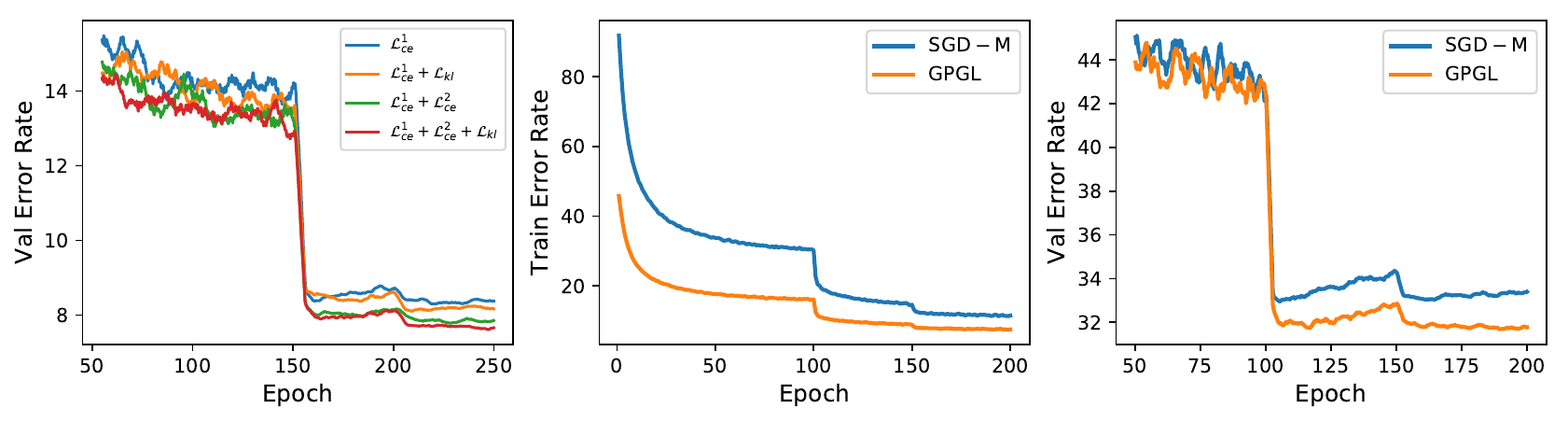}}
	\leftline{\small\hspace{0.5cm}(a) Triangle loss on CIFAR-10\hspace{1.0cm}(b) Train error on CIFAR-100\hspace{1.0cm}(c) Val error on CIFAR-100}
	\caption{(a) Different combinations of three loss terms effect the validation error on CIFAR-10, which shows the effectiveness of our triangle loss function. (b) The training error of Resnet20 on CIFAR-100 over epochs. (c) The validation error of Resnet20 on CIFAR-100 over epochs.  }
	\label{exp_example_2}
\end{figure*}

\paragraph{Validation of loss combinations}
With the constraints of two extra loss terms, we explore the contribution of each loss function in the deep learning process. We train ResNet20 on CIFAR-10 with 4 strategies: 1) $\mathcal{L}_{ce}^1$ only, 2) $\mathcal{L}_{ce}^1 + \mathcal{L}_{ce}^2$, 3) $\mathcal{L}_{ce}^1 + \mathcal{L}_{kl}$ and 4) triangle loss. In Figure~\ref{exp_example_2}(a), we can learn that the performance of triangle loss is better than the performance of using the combinations of 2 loss terms and these three strategies are all better than only using the $\mathcal{L}_{ce}^1$ loss. In other words, each of the two extra loss terms improve the performance and they can be used together.

\subsection{Performance comparison}
\label{Performance_comp}
In this part, we compare our GPGL scheme with a common SGD-Momentum optimizer (replicated by ourselves) in convergence speed on CIFAR-10, CIFAR-100 and Tiny-ImageNet. We report the performance and the number of epochs by which they reach the best performance in SGD-Momentum optimizer at the last error plateaus. The experiments are repeated 5 times and the results are shown in Table~\ref{Table_our_method}. 

From Table~\ref{Table_our_method}, we have the following observations. Firstly, our GPGL scheme achieves lower test errors and converges faster than SGD-Momentum for all the used CNN architectures on both three datasets. our GPGL scheme has an average improvement about 0.3$\%$ on CIFAR-10 and up to 1.47$\%$ on CIFAR-100 compared with SGD-Momentum. Our GPGL scheme uses about 50 to 90 less epochs to achieve the same accuracy as SGD-Momentum. This demonstrates the importance of the context labels, which can provide guidance to DNNs. 

We also compare performance of our approach with some state-of-the-art optimization strategies on CIFAR-10. The results are shown in Table~\ref{tab:table_SOTA} (notice that the models in BRLNN~\cite{lu18d} use the pre-activation strategy~\cite{he2016identity}), which shows that our GPGL scheme outperforms the other methods.


\begin{table}
	\caption{Comparison of SGD-M and GPGL}
	\begin{center} 
		\label{Table_our_method} 
	\begin{tabular}{lcccc}
		\toprule
		\multirow{2}*{\textbf{Model}} & \multicolumn{2}{c}{\textbf{SGD-M}} & \multicolumn{2}{c}{\textbf{GPGL}} \\
		\cmidrule(lr){2-3}\cmidrule(lr){4-5}
		& error (\%) & epochs & error (\%) & epochs \\
		\toprule
		\multicolumn{5}{c}{\textbf{CIFAR-10}} \\
		\toprule
		ResNet20 & 7.94 (8.18$\pm$0.14) & 231 & \textbf{7.67} (7.83$\pm$0.13)&158  \\ \hline
		ResNet32 & 6.94 (7.09$\pm$0.11) & 233 & \textbf{6.55} (6.88$\pm$0.18)&171  \\ \hline
		ResNet44 & 6.55 (6.76$\pm$0.14) & 242 & \textbf{6.23} (6.42$\pm$0.15)&163 \\ \hline
		ResNet56 & 6.20 (6.40$\pm$0.16) & 231 & \textbf{5.90} (6.13$\pm$0.20)&183 \\ \hline
		ResNet110 & 5.82 (5.97$\pm$0.18) & 235 & \textbf{5.48} (5.68$\pm$0.20)&173 \\ \hline
		PreActResNet20 & 7.80 (7.94$\pm$0.13)&228 & \textbf{7.53} (7.64$\pm$0.08)&169 \\ \hline
		ResNeXt29\_8\_64 & 4.18 (4.43$\pm$0.19)&236 & \textbf{4.11} (4.23$\pm$0.10)&178 \\
		\toprule
		\multicolumn{5}{c}{\textbf{CIFAR-100}} \\
		\toprule
		ResNet20 & 32.98 (33.16$\pm$0.14) & 164 & \textbf{31.63} (31.89$\pm$0.24) & 101\\ \hline
		ResNet110 & 27.81 (28.21$\pm$0.26) & 190 & \textbf{26.34} (27.07$\pm$0.39) & 101\\ \hline
		ResNeXt29\_8\_64 & 21.07 (21.32$\pm$0.16) & 186 & \textbf{20.58} (20.84$\pm$0.19) & 132 \\ \hline
		\toprule
		\multicolumn{5}{c}{\textbf{Tiny-ImageNet}} \\
		\toprule
		ResNet18 & 33.20 (33.58$\pm$0.24) & 115 & \textbf{32.36} (32.61$\pm$0.18) & 49 \\
		\bottomrule
	\end{tabular}
	\end{center}
\end{table} 


\begin{table}
	\caption{Comparison with state-of-the-art (error rate (\%))}
	\begin{center}
		\label{tab:table_SOTA} 
		\begin{tabular}{lcccc}
			\hline
			\toprule
			\textbf{Methods} & \textbf{SGD-M}~\cite{he2016deep} & \textbf{BFLNN}~\cite{lu18d} & \textbf{Others} & \textbf{Ours} \\ 
			\toprule
			ResNet20  & 8.75 & 8.33 & 7.89~\cite{han2016_DSD} & \textbf{7.67} \\ \hline
			ResNet32  & 7.51 & 7.18 & -    & \textbf{6.55} \\ \hline
			ResNet44  & 7.17 & 6.66 & -    & \textbf{6.23} \\ \hline
			ResNet56  & 6.97 & 6.31 & 6.86~\cite{hoffer2018fix} & \textbf{5.90} \\ \hline
			ResNet110 & 6.43 & 6.16 & 6.84~\cite{chen2018complement} & \textbf{5.48} \\
			\bottomrule
		\end{tabular}
	\end{center}
\end{table}

\section{Conclusion}
In this paper, we have presented an epoch-evolving Gaussian Process Guided Learning (GPGL) scheme to estimate the context-aware class distribution information and guide the conventional bottom-up learning process efficiently. 
We have demonstrated that our triangle consistency loss function is effective for a good balance between precise batch-level distribution learning and global distribution-aware nonparametric modeling. 
The experiments on CIFAR-10, CIFAR-100, and Tiny-ImageNet datasets have validated that our GPGL scheme improves the performance of CNN models remarkably and reduces the number of epochs significantly with better convergence performance.
\clearpage
\bibliographystyle{unsrtnat}     
\bibliography{mybibfile}
\end{document}